\documentclass[runningheads]{llncs}

\usepackage[T1]{fontenc}
\usepackage{graphicx}
\usepackage{tikz}
\usetikzlibrary{shapes,arrows}
\usepackage{graphicx}
\usepackage[section]{placeins}
\usepackage{booktabs}
\usepackage{makecell}
\usepackage{tabularx}
\usepackage{comment}
\usepackage{pgf}
\usepackage{subcaption}
\usepackage{comment}
\usepackage{xcolor}

\usepackage[ruled,vlined]{algorithm2e}

\usepackage{hyperref}

\usepackage{mwe}
\usepackage{tablefootnote}

\usepackage{algpseudocode}
\usepackage{booktabs}
\usepackage{multirow}
\usepackage{eurosym}
\usepackage{amssymb}
\usepackage{amsmath}
\usepackage{wasysym}
\usepackage{mathtools}
\usepackage{siunitx,etoolbox}
\usepackage{listings}
\usepackage{float} 
\usepackage{tabularx}

\usepackage{ragged2e}
\usepackage{array}
\usepackage{anyfontsize}
\usepackage{framed}
\usepackage[none]{hyphenat}
\usepackage{pgf}
\usepackage{enumitem} 

\usepackage{amssymb}
\allowdisplaybreaks

\usepackage{booktabs}
\begin{document}
\title{Leveraging Large Language Models to Develop Heuristics for Emerging Optimization Problems}
\titlerunning{Leveraging LLMs to Develop Heuristics for Emerging Optimization Problems}
%
\author{Thomas Bömer\inst{1}\orcidID{0000-0003-4979-7455} \and
Nico Koltermann\inst{1}\orcidID{0009-0008-0359-0452} \and
Max Disselnmeyer\inst{2}\orcidID{0009-0008-5689-2235}\and
Laura Dörr \inst{2}\orcidID{0000-0002-8007-1815}\and
Anne Meyer \inst{2}\orcidID{0000-0001-6380-1348}
}
\authorrunning{T. Bömer et al.}
%
\institute{TU Dortmund University, Leonhard-Euler-Stra{\ss}e 5, 44227 Dortmund, Germany \email{\{thomas.boemer,nico.koltermann\}@tu-dortmund.de} \and
Karlsruhe Institute of Technology, Kriegsstraße 77, 76133 Karlsruhe, Germany
\email{\{max.disselnmeyer,laura.doerr,anne.meyer\}}@kit.edu}
\maketitle              
\begin{abstract}
Combinatorial optimization problems often rely on heuristic algorithms to generate efficient solutions. However, the manual design of heuristics is resource-intensive and constrained by the designer's expertise. Recent advances in artificial intelligence, particularly large language models (LLMs), have demonstrated the potential to automate heuristic generation through evolutionary frameworks. 
Recent works focus only on well-known combinatorial optimization problems like the traveling salesman problem and online bin packing problem when designing constructive heuristics.
This study investigates whether LLMs can effectively generate heuristics for niche, not yet broadly researched optimization problems, using the unit-load pre-marshalling problem as an example case. 
We propose the Contextual Evolution of Heuristics (CEoH) framework, an extension of the Evolution of Heuristics (EoH) framework, which incorporates problem-specific descriptions to enhance in-context learning during heuristic generation. 
Through computational experiments, we evaluate CEoH and EoH and compare the results. 
Results indicate that CEoH enables smaller LLMs to generate high-quality heuristics more consistently and even outperform larger models. Larger models demonstrate robust performance with or without contextualized prompts. 
The generated heuristics exhibit scalability to diverse instance configurations.

\end{abstract}

\section{Introduction} 
\label{sec: Introduction}
Combinatorial optimization problems are pivotal in logistics, manufacturing, and supply chain management. Heuristic algorithms are commonly employed for such problems because they can provide good solutions quickly, making them suitable for real-world applications. However, designing heuristics traditionally requires significant human expertise and is time-consuming, making the process financially expensive.
\par
Automated heuristic design reduces reliance on specialized human expertise \cite{burke2013hyper}. Genetic programming is commonly used to evolve program code for solving optimization problems \cite{langdon2013foundations}. These methods are effective but rely on predefined search spaces. Human input determines permissible primitives and mutation operations for generating heuristics. This dependency may limit the discovery of novel approaches, especially in complex or less researched domains.
\par
Recent advancements in artificial intelligence have introduced large language models (LLMs) as a new resource for tackling optimization \cite{yang2024largelanguagemodelsoptimizersOPRO}. Though LLMs are primarily recognized for their natural language generation capabilities, studies show they can generate code and solve mathematical problems as well \cite{austin2021program,naveed2023comprehensive}. This broader skill set may help address the challenge of heuristic generation without relying on large, predefined search spaces.
\par
In combinatorial optimization, LLMs generally play two roles \cite{liu2024systematic}: \textit{optimizers} or \textit{designers}. Optimizer-based approaches request the LLM to solve the problem and return a solution \cite{yang2024largelanguagemodelsoptimizersOPRO,huang2024multimodalintegrationboostperformanceMLLM}, which shows moderate performance and little scalability. 
Designer-based approaches request LLMs to create heuristics, drawing on the models’ ability to generate and refine heuristics \cite{romera2024mathematical,liu2024evolutionEoH}. Although promising, most designer-based studies focus on well-known problems like the traveling salesperson problem (TSP) \cite{liu2024evolutionEoH,ye2024reevo} and online bin packing problem (oBPP) \cite{romera2024mathematical,liu2024evolutionEoH,dat2024hsevo} when designing constructive heuristics. Their usage in niche optimization problems remains unexplored.
\par
This work investigates whether LLMs can generate effective constructive heuristics for a specialized yet practically important niche optimization problem: the unit-load pre-marshalling problem (UPMP) \cite{pfrommer2023solving}. 
The UPMP arises in block-stacking warehouses where unit loads must be rearranged to remove blockages that block higher-priority loads. It requires a structured procedure to select moves to reach a blockage-free state.
We embed the generated heuristic in a tree search framework.
First, each possible reshuffling move is branched from an initial warehouse state. Then, the LLM-generated heuristic rates each resulting warehouse state. The highest-rated state is selected to proceed with the next iteration. This cycle repeats until a blockage-free state is achieved.
\par 
We find the UPMP a strong candidate for testing LLM-based heuristic design in niche optimization problems for the following reasons: 
The problem description is most likely less prevalent in the training data of LLMs. 
Very few or no heuristics code bases exist online on which an LLM may be trained. 
Related problems - such as the container pre-marshalling problem (CPMP) \cite{jimenez2023constraint} and block relocation problem \cite{kang2006planning} - share similar constraints, but are not nearly as widely studied as better-known problems like TSP or oBPP.
\par
We introduce the Contextual Evolution of Heuristics (CEoH) framework, an extension of the Evolution of Heuristics framework (EoH) \cite{liu2024evolutionEoH} that incorporates problem-specific details into the heuristics design process. By adding a “contextual” component to each prompt, CEoH enables LLMs to better understand the problem and its constraints. The result is a more informed heuristic-generation process, where the model uses in-context learning to generate heuristics tailored to the UPMP.
\par
The rest of the paper is organized as follows. Section \ref{sec: The Unit-load Pre-marshalling Problem} defines the UPMP. Section \ref{sec: related work} reviews relevant research on pre-marshalling and LLM-based heuristic design. Section \ref{sec: Contextual Evolution of Heuristics Framework} introduces the CEoH framework. Section \ref{sec: Computational Experiments} presents the results of computational experiments. Finally, Section \ref{sec: Conclusion} summarizes the findings and outlines future research directions.

\section{The Unit-load Pre-marshalling Problem}
\label{sec: The Unit-load Pre-marshalling Problem}
The UPMP, introduced by \cite{pfrommer2023solving}, is about sorting unit loads using autonomous mobile robots (AMR) in a block-stacking warehouse according to predefined priorities to remove blockages. 
A block-stacking warehouse is a storage system in which unit loads such as pallets are stacked without additional infrastructure.
During pre-marshalling, no unit load enters or leaves the warehouse. 
A unit load is deemed \textit{blocking} if it hinders access to a unit load with a higher priority class. 
Priority classes can, for example, be assigned based on the expected retrieval time.

\begin{figure}[!ht]
    \centering
    \includegraphics[width=0.75\textwidth]{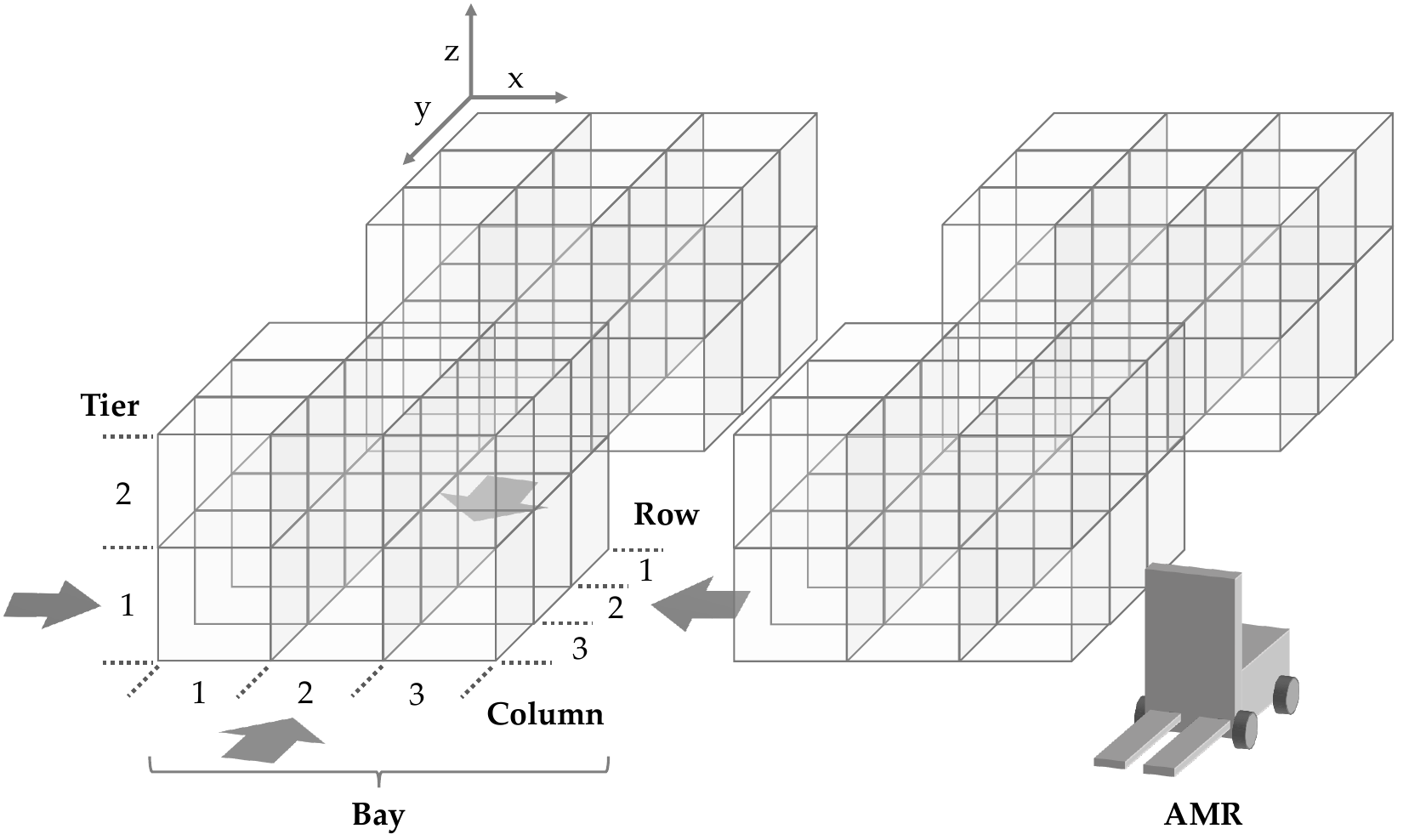}
    \caption{Representation of a two-tier multibay block-stacking warehouse with three rows and columns per bay and four bays in total \cite{pfrommerboemer2024sorting}. This configuration displays a 3x3 bay layout and 2x2 warehouse layout.
    } 
    \label{fig:multi_block_storage_example}
\end{figure}

The warehouse consists of a set of interconnected bays $\mathcal{B}$. Each bay $b \in \{1,...,|\mathcal{B}|\}$ can be accessed from up to four access directions. Unit loads are placed inside a bay in a shared storage policy. No empty gaps inside the bay are allowed.
Figure \ref{fig:multi_block_storage_example} illustrates the configuration of a block-stacking warehouse consisting of four bays arranged in a square layout. Let $\mathcal{C}$ be the set of columns, $\mathcal{R}$ the set of rows, and $\mathcal{T}$ the set of tiers.
Each bay features a three-dimensional grid of storage slots. Each storage slot is denoted by column $c \in\{1,...,|\mathcal{C}|\}$, row $r \in \{1,...,|\mathcal{R}|\}$, and tier $t \in \{1,...,|\mathcal{T}|\}$. 
Let $\mathcal{P}$ be the set of priority classes. The grid is filled with unit loads categorized into priority classes. The priority class in a grid position is denoted by $p_{crt} \in \{1,...,|\mathcal{P}|\}$. 
Let $\mathcal{A}$ be the set of access points. An access point $a \in \{1,...,|\mathcal{A}|\}$, is located in the aisle space in front of the bay, determining the slots that can be accessed from that point. 
Slots that are accessed from the same access point are called lanes.
Only the topmost, outermost unit load of a lane is accessible.
\par
In this study, we focus on the UPMP with only one-tier and one access direction. \cite{pfrommer2023solving} find the one access direction variant is the most challenging because blockages cannot be avoided by selecting favorable access directions for each unit load. 
\par
\begin{figure}
    \centering
    \includegraphics[width=1\linewidth]{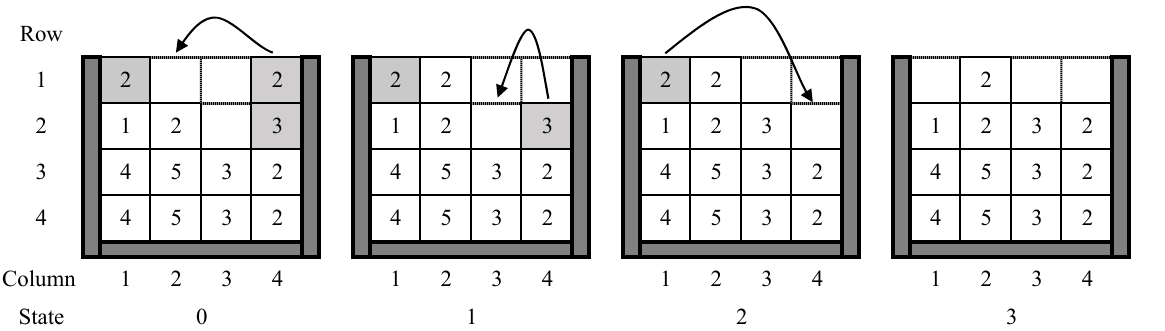}
    \caption{Example sequence of moves to solve an instance. Top-down view on a single-bay. Unit loads can be accessed from the north direction only. }
    \label{fig:instance_north_access_example}
\end{figure}

Figure \ref{fig:instance_north_access_example} shows a one-tier single-bay warehouse problem instance from a top-down perspective. The unit loads can only be accessed from the north. In the initial state 0, three unit loads are marked as blocking. They block access to higher-priority class unit loads or need to be removed to resolve a blockage. All blockages are cleared after three moves (state 3).

\section{Related Work}
\label{sec: related work}
In this Section, we first present related work on pre-marshalling problems. Following this, we present studies employing LLMs in an evolutionary framework to develop heuristics for combinatorial optimization problems.

\subsection{Related Work on Pre-marshalling}
For the UPMP in single-bay block-stacking warehouses, \cite{pfrommer2023solving} introduce a two-step approach. First, the authors use a network flow formulation to decompose the warehouse bays into virtual lanes. Afterward, an A*-algorithm-based tree search solves the problem with minimizing the number of moves.
The approach in \cite{pfrommer2023solving} for a single-bay warehouse is extended by the authors of \cite{pfrommerboemer2024sorting} to a multibay warehouse. The approach minimizes the loaded move distance for the minimal number of reshuffling moves. 
\cite{bomer2024sorting} extend the approach of \cite{pfrommerboemer2024sorting} by proposing a sequential approach that assigns the moves found by the tree search to robots according to move dependencies using a mixed-integer formulation.
\par
The pre-marshalling of unit loads in a block-stacking warehouse is closely related to the CPMP in harbor logistics. The CPMP aims to rearrange stacks of containers within a bay until no blockages persist. 
Solution approaches exist in the form of greedy heuristics \cite{exposito2012pre}, tree search approaches \cite{bortfeldt2012tree,wang2024policy,hottung2020deep}, 
dynamic programming \cite{prandtstetter2013dynamic}, 
integer programming \cite{lee2007optimization,parreno2019integer},
and constraint programming 
\cite{rendl2013constraint,jimenez2023constraint}. 
Since cranes in maritime harbors can often only operate within one bay, no adaptation of the CPMP considers multiple blocks (here: bays).
\par
For the UPMP and CPMP heuristic and exact approaches have been developed to minimize the number of moves. However, the design required extensive algorithmic and domain knowledge.

\subsection{Application of LLMs in Combinatorial Optimization}
In this section, we present studies that use LLMs to generate constructive heuristics for combinatorial optimization problems in an evolutionary framework.
\par
FunSearch, proposed in \cite{romera2024mathematical}, is the first approach that pairs a pre-trained LLM with an evolutionary procedure to explore algorithms. FunSearch starts with a population of manually designed very basic, low-scoring heuristics. Each heuristic is represented by program code. An LLM is requested to generate a new heuristic based on parent heuristics. The generated heuristic is evaluated on test instances and saved to a database. The heuristics are represented as program code.
\par
\par
The authors of \cite{liu2024evolutionEoH} introduce Evolution of Heuristics (EoH) as an extension of Funsearch \cite{romera2024mathematical}.
In EoH, heuristic ideas are represented as natural language descriptions called 'thoughts' and program code. This approach evolves thoughts and program code simultaneously, leveraging the models to generate and refine heuristics effectively. 
Further, EoH introduces two exploration and three modification prompt strategies.
The results show that EoH outperforms handcrafted heuristics and FunSearch, while using significantly fewer LLM queries (around one million queries in FunSearch vs. 2000 queries EoH for oBPP). 
\par
The authors of \cite{ye2024reevo} propose the Reflective Evolution (ReEvo) framework. ReEvo uses evolutionary search combined with model-generated reflections to explore and modify heuristics more efficiently. 
The iterative process includes the following five steps: selection, short-term reflection, crossover, long-term reflection, and elitist mutation.
Selection selects random parent heuristics; the short-term reflection analyzes the selected heuristics and provides hints for evolutionary search; the LLM performs a crossover mutation by generating a new heuristic based on the parent heuristics, their relative performance, and reflector output; the long-term reflection accumulates expertise among short term reflections; finally in the elitist mutation an LLM samples multiple heuristics based on the current best heuristic and the long-term reflection.
\par
\par
The latest approach we consider, HSEvo \cite{dat2024hsevo}, aims to balance diversity and optimization performance by leveraging harmony search alongside traditional evolutionary operators. 
The framework introduces two diversity measurement metrics 
to quantify heuristic exploration and ensure diversity in the search space. 
HSEvo employs a multi-stage optimization pipeline, including flash reflection, an efficient alternative to ReEvo’s reflection mechanism, elitist mutation to refine top-performing heuristics, and harmony search to fine-tune heuristic parameters. 
Experimental results demonstrate that HSEvo achieves superior heuristic diversity and optimization performance, outperforming FunSearch, EoH, and ReEvo in both solution quality and computational efficiency.
\par
\par
These studies demonstrate the potential of LLMs in generating constructive heuristics for combinatorial optimization problems.
However, the approaches focus on well-established optimization problems as summarized in Table \ref{tab:problem_view_optimization}, leaving niche problems under-researched. 
It is unclear whether LLMs can generate good constructive heuristics for problems for which they have less contextual knowledge.
This study aims to bridge this gap by applying LLM-generated constructive heuristics to the UPMP.

\begin{table}[htpb]
\caption{Combinatorial optimization problems addressed with the design of a constructive heuristic in related papers.}
\centering
\begin{tabular}{llccc}
\toprule
Paper & Framework & TSP & oBPP  & UPMP
\\
\midrule
\cite{romera2024mathematical} & FunSearch & & \checkmark & \\
\cite{liu2024evolutionEoH} & EoH & \checkmark & \checkmark & \\
\cite{ye2024reevo} & ReEvo & \checkmark & & \\
\cite{dat2024hsevo} & HSEvo & & \checkmark & \\
 & CEoH & & & \checkmark\\
\bottomrule
\end{tabular}
\label{tab:problem_view_optimization}
\end{table}

\section{Contextual Evolution of Heuristics Framework}
\label{sec: Contextual Evolution of Heuristics Framework}
\begin{figure}
    \centering
    \includegraphics[width=1\linewidth]{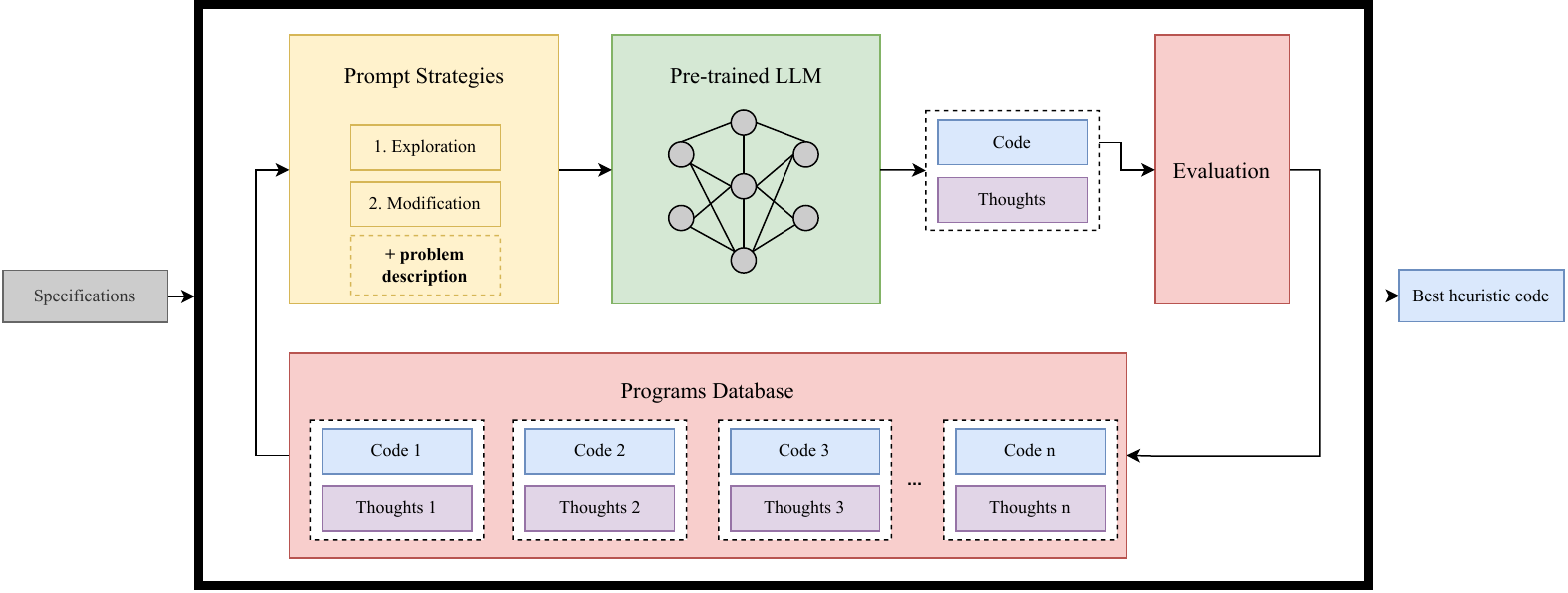}
    \caption{The CEoH framework evolves code and thoughts while using an additional problem description.}
    \label{fig:CEoH_concept}
\end{figure}

This section describes the Contextual Evolution of Heuristics (CEoH).
The core idea of CEoH is to extend EoH \cite{liu2024evolutionEoH} with an additional optimization problem description in the prompt to leverage in-context learning.
Figure \ref{fig:CEoH_concept} shows the concept of CEoH. 

\paragraph{Evolutionary Procedure.}
CEoH is an iterative evolutionary framework that generates and evolves heuristics. Each \textit{heuristic} $h \in \mathcal{H}= \{1, 2, \dots, |\mathcal{H}|\}$ is represented by program code and \textit{thoughts}. 
The program code is a Python function with a defined input and output.
The thoughts describe the core ideas the LLM had while generating the heuristic.
\par
The approach iterates a set of generations $\mathcal{G} = \{1,2,\dots,|\mathcal{G}|\}$. 
The population of heuristics in generation $g \in \mathcal{G}$ is denoted as $\mathcal{P}_{g} = \{1, 2, \dots, |\mathcal{P}_{g}|\}$. 
In each generation $g$, a set of prompt strategies $\mathcal{S} = \{1,2,\dots,|\mathcal{S}|\}$ is iterated to generate the population of heuristics $\mathcal{P}_{g}$ based on parent heuristics using a pre-trained LLM. Each prompt strategy $s \in \mathcal{S}$ is used $\bar{r}$ times. Hence, $ |\mathcal{S}| \cdot \bar{r}$ heuristics are generated each generation $g$ for the population $\mathcal{P}_{g}$.
Each new heuristic $h$ is evaluated on a set of problem instances $\mathcal{I} = \{1,2,\dots,|\mathcal{I}|\}$ and assigned a fitness value $f^{\mathcal{I}}(h)$. Then, the heuristic $h$ is added to the current population $\mathcal{P}_{g}$ and the next heuristic is prompted.
After a full generation ($ |\mathcal{S}| \cdot \bar{r}$ prompts), the current population's best $\bar{n}$ heuristics are taken to the next generation. 
Before the iterative procedure, an initialization prompt is called $2 \cdot \bar{n}$  times to generate $\mathcal{P}_{0}$.

\paragraph{Prompt Strategies}
For the prompt strategies, we adopt the initialization prompt I0, the exploration strategy prompts E1 and E2, and the modification strategy prompts M1 and M2 from \cite{liu2024evolutionEoH}. We briefly describe the prompt strategies for completeness. 

\begin{itemize}[label={}]
  \item \textbf{I0:} Generate a heuristic to solve the optimization problem.
  \item \textbf{E1:} Generate a completely different heuristic based on $\bar{p}$ parent heuristics selected from the current population.
  \item \textbf{E2:} Generate a new heuristic inspired by the same core idea as $\bar{p}$ parent heuristics selected from the current population.
  \item \textbf{M1:} Modify a parent heuristic from the current population for better performance.
  \item \textbf{M2:} Modify the parametrization of a parent heuristic from the current population for better performance.
\end{itemize}

In EoH \cite{liu2024evolutionEoH} the authors only provide a very brief task description. 
This may be sufficient for well-known problems like the oBPP and TSP. 
However, we argue that additional problem description helps to design heuristics in less-known niche optimization problems.
Hence, in contrast to EoH we enrich the prompt with the additional problem description. 
\par
Each prompt follows the same structure: (1) task description, (2) additional problem description (new in CEoH), (3) parent heuristic(s) (not in I0), (4) strategy-specific output instructions, and (5) additional instructions. We detail each component in the following:
The (1) task description informs the LLM about the optimization problem and how the heuristic will be used in a bigger context - for example, to score warehouse states in a tree search. 
The (2) additional problem description enhances in-context learning by specifying the expected input's data structure and clarifying how it should be interpreted within the problem domain. This section also includes examples of program input and output to illustrate expected behavior and explicit instructions requiring the model to thoroughly analyze the additional problem description.
The inclusion of (3) parent heuristic(s) in the form of program code and corresponding thoughts facilitates few-shot learning, ensuring the LLM can generalize from prior examples.
The (4) strategy-specific output instructions guide the generation process by incorporating a prompt-strategy-specific sentence, instructions to formulate the thoughts behind a new heuristic, and a program code for that heuristic with defined input and output.
(5) Additional instructions impose key implementation constraints. These include specifying input and output data types while discouraging nested methods to minimize coding errors. To improve solution robustness, the use of random components is restricted. Self-consistency is reinforced to ensure that the generated Python function adheres to previously formulated thoughts \cite{min2023beyond}. Finally, the LLM is instructed not to provide additional explanations to optimize prompt efficiency by reducing generated tokens.

\paragraph{Fitness function.}
Each heuristic $h$ has to be evaluated for its fitness. 
Let $\mathcal{I}$ be the set of problem instances. 
Let $m_{i}$ be the number of moves the heuristic $h$ needs to solve problem instance $i \in \mathcal{I}$. 
Let $m^{lb}_{i}$ be the lower bound of moves to solve the problem instance $i$. 
We obtain the lower bound $m^{lb}_{i}$ according to the approach of \cite{pfrommer2023solving}. 
We assume a heuristic to be unable to solve an instance $i$ if a solution was not found after a maximum move number $m^{max}$. $m^{max}$ is selected significantly larger than $m^{lb}_{i}$. We set $m_{i} = m^{max}$, if $m^{max}$ is reached in the evaluation thereby effectively limiting the maximum fitness value of a heuristic.
This fitness function design promotes heuristics that are able to solve a wide range of problem instances. Hence, we first search for a heuristic that can solve all problem instances and then for a heuristic that solves all problem instances efficiently with minimal moves. 
The Equation \ref{eq:fitness_funtion} calculates the fitness as the average relative difference between the heuristic solution $m_{i}$ and the lower bound $m^{lb}_{i}$. 
We seek to minimize the fitness function value.
A heuristic $h$ with $f^{\mathcal{I}}(h)=0.2$ requires on average 20\,\% moves more on the problem instances $\mathcal{I}$ than the lower bound.
Please note that the lower bound $m^{lb}_{i}$ is not necessarily equal to the optimal solution. Hence, a fitness of zero may not be achievable.

\begin{equation}
\label{eq:fitness_funtion}
f^{\mathcal{I}}(h) = 
\frac
{
1
}
{
|\mathcal{I}|
}
\displaystyle\sum_{i \in \mathcal{I}}\frac{m_{i}-m^{lb}_{i}}{m^{lb}_{i}}
\end{equation}

\section{Computational Experiments}
\label{sec: Computational Experiments}
We conducted extensive experiments 
\footnote{The  source code to reproduce the experiments can be found in:\newline  
https://github.com/nico-koltermann/contextual-evolution-of-heuristics}
to evaluate the effect of CEoH introduced in Section \ref{sec: Contextual Evolution of Heuristics Framework}.
We apply the CEoH and EoH \cite{liu2024evolutionEoH} framework to the UPMP and compare the generated heuristics.


\paragraph{Parameters.}
We adapted the same parameters for CEoH and EoH.
As pre-trained LLMs we consider the open-source models: \texttt{Gemma2:27b} \cite{gemma2_27b}, \texttt{Qwen2.5-Coder:32b} \cite{qwen2.5-coder_32b}, and \texttt{DeepseekV3:685b} \cite{deepseekV3} and the closed-source models: \texttt{GPT4o:2024-08-06} and \texttt{GPT4o:2024-11-20} \cite{gpt4o}.
We select different LLMs to observe the effect of context size and coding focus.
Among these, \texttt{Qwen2.5-Coder:32b} and \texttt{DeepseekV3:685b} are explicitly optimized for code-related tasks, incorporating specialized pretraining on programming languages and structured reasoning. In contrast, the \texttt{GPT4o} LLMs are general-purpose but have demonstrated strong coding capabilities.
All LLM sampling hyperparameters, such as temperature and top-p, are at their default value to limit the design space of our study.
We execute 10 experiment runs (see Section \ref{sec: Contextual Evolution of Heuristics Framework} - Evolutionary Procedure) for CEoH and EoH with each LLM. 
The number of generations is set to 20.
The $\bar{n} = 20$ best heuristics are taken to the next generation.
We use the prompt strategy I0 for the initialization and the four prompt strategies E1, E2, M1, and M2 to evolve the heuristics.
The number of parent heuristics for E1 and E2 is set to $p = 5$.
The initialization prompt I0 is called $40$ times. The $20$ best heuristics form the initial population.
The other prompt strategies (E1, E2, M1, and M2) are each called $\bar{r} = 20$ times per generation.
Hence, we perform $20 \cdot 4 \cdot 20 = 1600$ prompts in one experimental run to evolve the initial heuristics.
\par
The evaluation instances have an average lower bound of $11.8$ moves.
Hence, for the evaluation of heuristics, we set a sufficiently large maximum move limit of $m^{max}=100$ moves to penalize unsolved instances. 
The configuration of the evaluation instances is: 5x5 bay layout, 1x1 warehouse layout, one access direction (north), one-tier, five priority classes, and 60\,\% fill percentage. 
A 5x5 bay layout describes bays with five rows and five columns - 25 slots for the one-tier case. 60\,\% of these slots are filled with priority classes of one (highest) to five (lowest).
A 1x1 warehouse layout denotes a warehouse that consists of a single bay. 
One-tier means that no stacking is performed.
We consider 10 instances (seeds 0-9) for the evaluation. 
All heuristics were evaluated on a machine with an AMD EPYC 7401P processor and 64 gigabytes of RAM.

\begin{figure}[h!t]
    \centering
    \includegraphics[width=\linewidth]{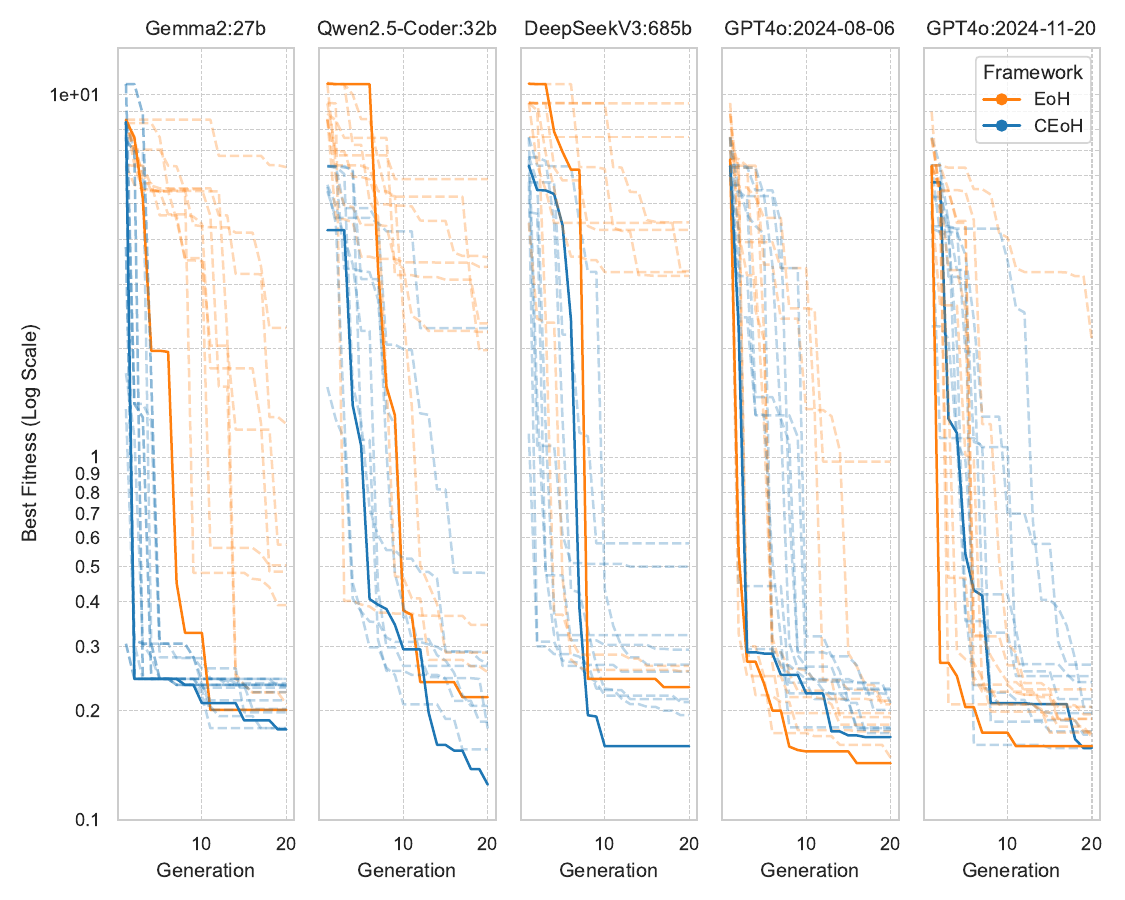}
    \caption{Best found heuristic across generations for each model and experiment run. The best run for CEoH and EoH is shown with opacity. Lower values indicate better performance.}
    \label{fig:best-heuristic-models}
\end{figure}

\paragraph{Fitness.}
Figure \ref{fig:best-heuristic-models} shows the fitness of the best heuristics found up to each generation and experiment run. The opaque lines represent the experiment runs that yielded the best heuristics in the final population for each framework.

CEoH demonstrates superior performance for the models: \texttt{Gemma2:27b}, \texttt{Qwen2.5-Coder:32b}, and \texttt{DeepSeekV3:685b}. 
The additional problem context in CEoH enables smaller models, especially, to generate overall better heuristics and to generate good heuristics more robustly.
However, for the large models \texttt{GPT4o:2024-08-06} and \texttt{GPT4o:2024-11-20} CEoH and EoH reliably found heuristics with a fitness below 27\,\% except for two EoH outlier runs. 
For the \texttt{GPT4o:2024-08-06} CEoH generally generates worse heuristics than EoH.
This suggests that the added problem description in CEoH may have restricted the model’s ability to leverage its pre-existing optimization knowledge effectively.
\par
The overall best heuristic in all experiment runs is generated by \texttt{Qwen2.5-Coder:32b} in the CEoH framework with a fitness of 12.5\,\%. The best heuristic for the EoH framework is generated by \texttt{GPT4o:2024-08-06} with a fitness of 14.3\,\%.

\paragraph{Generated heuristic code.}
Figure \ref{fig:code-examples} compares the best-scored heuristics generated by (a) \texttt{Qwen2.5-Coder:32b} CEoH and (b) \texttt{GPT4o:2024-08-06} EoH.
\par
(a) The \texttt{Qwen2.5-Coder:32b} CEoH heuristic evaluates lanes based on priority balance, blocking penalties, and density weighting. It rewards sequences where unit loads follow a decreasing priority order and penalizes configurations that create blockages. The heuristic also incorporates a dynamic penalty scaling mechanism, where higher-density lanes and misplaced units receive greater penalties. Additionally, non-blocking lanes gain extra bonuses.
\par
(b) The \texttt{GPT4o:2024-08-06} EoH heuristic employs a logistic transformation approach to dynamically adjust scores based on priority values and lane positioning. It applies exponential penalties to blocking unit loads while rewarding accessible, well-ordered stacks. 
\par
Notably, (b) \texttt{GPT4o:2024-08-06} EoH refers to lanes as "stacks", a term associated with container pre-marshalling. We do not use the word "stack" in any prompt. We hypothesize, this suggests that \texttt{GPT4o:2024-08-06} EoH drew from related optimization problems when designing its heuristic, whereas \texttt{Qwen2.5-Coder:32b} CEoH heuristic relied more on the explicit problem context provided by CEoH.
In all \texttt{GPT4o:2024-08-06} experiment runs, the term "stack" occurred 86,291 times within the EoH framework and 63 times within the CEoH framework.
For the same model, the term "lane" appeared 22 times in heuristics generated within the EoH framework and 121,059 times in heuristics generated within the CEoH framework.
The other LLMs show a similar behavior as \texttt{GPT4o:2024-08-06}.

\lstset{
    language=Python,
    basicstyle=\ttfamily\tiny,
    keywordstyle=\bfseries\color{blue},
    commentstyle=\itshape\color{gray},
    stringstyle=\color{red},
    numbers=left,
    numberstyle=\tiny,
    stepnumber=1,
    frame=single,
    tabsize=2,
    showstringspaces=false,
    xleftmargin=0pt,
    xrightmargin=0pt,
    aboveskip=1em,  
    belowskip=1em   
}

\begin{figure}[ht]
\centering

\begin{subfigure}{0.45\textwidth}
\begin{lstlisting}
"""
The new algorithm scores each warehouse state by 
evaluating the accessibility of high-priority 
items with penalties for blocking units that are 
more severe if they block higher priority items, 
rewards for lanes with minimal blocking, and 
incorporates a novel scoring mechanism using 
linear decay for blocked items and priority-based 
bonuses adjusted by lane density with modified 
parameters.
"""

def select_next_move(warehouse_states):
  scores = []
  for state in warehouse_states:
    score = 0
    total_units = sum(len([unit for unit
                           in lane if unit != 0])
                      for lane in state)
    num_lanes = len(state)

    for i, lane in enumerate(state):
      highest_priority_seen = float('inf')
      blocking_occurred = False
      non_zero_count = len([unit for unit
                            in lane if unit != 0])
      density_weight = ((non_zero_count
                         / total_units)
                        ** 2 if total_units
                                > 0 else 1)
      block_penalty_factor = (4 + density_weight
                              * num_lanes * 1.2)
      priority_balance = sum(unit * (5 - unit)
                             for unit in lane)

      for j, unit in enumerate(reversed(lane)):
        if unit != 0:
          if unit > highest_priority_seen:
            penalty = ((unit ** 2)
                       * block_penalty_factor
                       * (0.9 ** j))
            score -= penalty
            blocking_occurred = True
          else:
            highest_priority_seen = unit

      score += (priority_balance * density_weight
                * 1.8)
      if not blocking_occurred:
        score += (non_zero_count ** 2
                  * (1 + density_weight * 0.6))

    scores.append(score)
  return scores
\end{lstlisting}
\caption{Qwen2.5:32b-Coder CEoH}
\end{subfigure}
\hfill
\begin{subfigure}{0.45\textwidth}
\begin{lstlisting}
"""
This heuristic implements an exponential function 
to further penalize blocked high-priority loads 
and further rewards states where high-priority 
loads are earlier in the stack to prioritize
reshuffling moves that release high-priority 
loads more efficiently.
"""

import math

def select_next_move(warehouse_states):
  scores = []
  for state in warehouse_states:
    score = 0
    for stack in state:
      bonus = 0
      penalty = 0
      can_access = True
      for i in range(len(stack) - 1, -1, -1):
        priority_adjustment =\
            1 / (1 + math.exp(-0.7 * stack[i]))
        if can_access:
          bonus += (stack[i] * 
                    (1 + math.exp(-0.5 * i)))
        if i > 0 and stack[i] < stack[i - 1]:
          penalty += ((1 - priority_adjustment) *
                      (stack[i - 1] - stack[i]) 
          * (1 / (1 + math.exp(0.5 * i))))
          can_access = False
      score += bonus - penalty
    scores.append(score)
  return scores
\end{lstlisting}
\caption{GPT4o:2024-08-06 EoH}
\end{subfigure}

\caption{Heuristic thoughts and code with best fitness value for CEoH and EoH.}
\label{fig:code-examples}
\end{figure}

\paragraph{Further Problem Instances.} To assess the applicability of the generated heuristics to various instance configurations, we evaluated the best-performing heuristic from each framework (see Figure \ref{fig:code-examples}) on additional problem instances. The performance was also compared with the state-of-the-art optimal A* approach for the UPMP proposed by \cite{pfrommer2023solving}.
\par
We vary the problem instance parameters bay layout, warehouse layout, and fill percentage.
Table \ref{tab:mean_lower_bound} shows the mean move lower bound \cite{pfrommer2023solving} for the evaluated instances for 60\,\%  and 80\,\% fill percentage.
An increase in fill percentage, bay layout size, and warehouse layout size causes additional moves. 
The largest instances need on average at least $157.53$ moves to be solved.
Considering this, we set the maximum move limit to $m^{max}=1000$ moves to ensure feasibility for all instance configurations.
While the warehouse layout size especially contributes to the number of moves, the fill percentage and bay layout size affect the complexity of the problem instance. 
A higher fill percentage reduces the repositioning options while maintaining the same number of slots.
The larger bay layout size causes deeper blockages which are harder to resolve. 

\begin{table}[h!t]
\centering
\caption{Mean move number lower bound for different instance parameters (seed 10-19).}
\label{tab:mean_lower_bound}

\begin{subtable}{0.49\textwidth}
\centering
\caption{60\,\% fill percentage}
\begin{tabular}{@{\extracolsep{3pt}}lrrr@{}}
\toprule
&\multicolumn{3}{l}{warehouse} \\
\cmidrule{2-4}
bay & 1x1 & 2x2 & 3x3 \\
\midrule
4x4 & 3.85 & 7.05 & 12.11 \\
5x5 & 12.17 & 26.28 & 42.94 \\
6x6 & 27.61 & 57.44 & 97.24 \\
\bottomrule
\end{tabular}
\end{subtable}
\hfill
\begin{subtable}{0.49\textwidth}
\centering
\caption{80\,\% fill percentage}
\begin{tabular}{@{\extracolsep{3pt}}lrrr@{}}
\toprule
&\multicolumn{3}{l}{warehouse} \\
\cmidrule{2-4}
bay & 1x1 & 2x2 & 3x3 \\
\midrule
4x4 & 6.67 & 13.11 & 19.63 \\
5x5 & 23.95 & 41.95 & 71.37 \\
6x6 & 50.47 & 95.94 & 157.53  \\
\bottomrule
\end{tabular}
\end{subtable}
\end{table}

Table \ref{tab:performance_overview} summarizes the performance of the best \texttt{Qwen2.5:32b-Coder} CEoH heuristic and best  \texttt{GPT4o:2024-08-06} EoH heuristic across various problem parameterizations. The top row displays the performance for the evaluation instances used in the evolutionary process (seed 0–9). Both heuristics perform very well on these evaluation instances. 
It is important to note that the fitness for A* is not necessarily zero, since the fitness is computed relative to the lower bound of moves. 
For example, the A*'s 8.15\,\% fitness value for a 60\,\% fill percentage, 5x5 bay layout, and 1x1 warehouse layout (seed 0-9) indicates that the optimal solution has on average 8.15\,\% more moves than the lower bound, highlighting a $12.5\,\% - 8.15\,\% = 4.35\,\%$ optimality gap for the best \texttt{Qwen2.5:32b-Coder} CEoH heuristic and a 6.15\,\%  optimality gap for the \texttt{GPT4o:2024-08-06} EoH heuristic.
\par
A comparison of ten more instances (seed 10-19) with the same configurations as the evaluation instances (60\,\% fill percentage, 5x5 bay layout, and 1x1 warehouse layout) shows that the \texttt{Qwen2.5:32b-Coder} CEoH heuristic still outperforms the \texttt{GPT4o:2024-08-06} EoH heuristic. This highlights the superiority of the \texttt{Qwen2.5:32b-Coder} CEoH heuristic for the evaluation configuration. However, the fitness for the new instances (seed 10-19) is significantly higher than for the evaluation instances (seed 0-9). This indicates an over-fitting to the training instances.

\begin{table}[h!t]
\caption{Performance overview for the best heuristics and the state-of-the-art optimal approach. The fitness [\%] and mean evaluation runtime [s] (time) are reported for the number of solved instances (sol.).}
\label{tab:performance_overview}
\begin{tabular*}{\linewidth}
{@{\extracolsep{\fill}}lllccccccccc}
\toprule
 &  &  & \multicolumn{3}{c}{Qwen2.5-Coder:32b CEoH} & \multicolumn{3}{c}{GPT4o:2024-08-06 EoH} & \multicolumn{3}{c}{A* \cite{pfrommer2023solving}} \\
\cmidrule{4-6} \cmidrule{7-9} \cmidrule{10-12} 
fill & bay & wh & sol. & fitness  & time & sol. & fitness  & time  & sol. & fitness& time \\
\midrule
\multicolumn{12}{c}{seed 0 - 9}\\
\midrule
60\% & 5x5 & 1x1 & 10 & 12.5 & 0.014 & 10 & 14.3 & 0.013 & 10 & 8.1 & 0.277 \\
\midrule
\multicolumn{12}{c}{seed 10 - 19}\\
\midrule
\multirow[t]{9}{*}{60\% } & \multirow[t]{3}{*}{4x4} & 1x1 & 9 & 23.5 & 0.005 & 7 & 12.4 & 0.003 & 10 & 3.3 & 0.064 \\
 &  & 2x2 & 10 & 16.5 & 0.534 & 10 & 2.6 & 0.770 & 9 & 0 & 1.166 \\
 &  & 3x3 & 10 & 19.7 & 12.25 & 10 & 2.9 & 19.33 & 8 & 0 & 10.91 \\
\cmidrule{2-12}
 & \multirow[t]{3}{*}{5x5} & 1x1 & 10 & 30.8 & 0.010 & 10 & 33.9 & 0.015 & 10 & 12.1 & 0.440 \\
 &  & 2x2 & 10 & 23.2 & 2.277 & 10 & 4.5 & 2.997 & 5 & 0 & 3.739 \\
 &  & 3x3 & 10 & 27 & 50.49 & 10 & 3.1 & 77.69 & 1 & 0 & 130.3 \\
\cmidrule{2-12}
 & \multirow[t]{3}{*}{6x6} & 1x1 & 10 & 32.5 & 0.040 & 9 & 27.1 & 0.050 & 9 & 6.9 & 1.944 \\
 &  & 2x2 & 10 & 30.7 & 5.865 & 10 & 4.6 & 10.63 & 4 & 0 & 28.06 \\
 &  & 3x3 & 10 & 48.5 & 156.9 & 10 & 3.1 & 244.9 & 1 & 0 & 211.1 \\
\cmidrule{1-12}
\multirow[t]{9}{*}{80\% } & \multirow[t]{3}{*}{4x4} & 1x1 & 1 & 0 & 0.001 & 1 & 50 & 0.003 & 5 & 50 & 0.250 \\
 &  & 2x2 & 10 & 39.8 & 0.631 & 10 & 18.6 & 0.916 & 4 & 1.2 & 51.27 \\
 &  & 3x3 & 10 & 37.8 & 12.55 & 10 & 6.3 & 22.64 & 1 & 0 & 170.1 \\
\cmidrule{2-12}
 & \multirow[t]{3}{*}{5x5} & 1x1 & 0 & - & - & 0 & - & - & 3 & 19.9 & 10.30 \\
 &  & 2x2 & 10 & 47.3 & 2.661 & 10 & 19.4 & 4.046 & 1 & 0 & 12.51 \\
 &  & 3x3 & 10 & 42.6 & 50.83 & 10 & 10.1 & 87.37 & - & - & - \\
\cmidrule{2-12}
 & \multirow[t]{3}{*}{6x6} & 1x1 & 0 & - & - & 0 & - & - & 1 & 29.4 & 581.4 \\
 &  & 2x2 & 10 & 50.5 & 8.949 & 10 & 23.1 & 15.33 & 0 & - & - \\
 &  & 3x3 & 10 & 53.7 & 162.2 & 10 & 18.9 & 258.9 & 0 & - & - \\
\bottomrule
\end{tabular*}
\end{table}
\par
When applying the heuristics to other instance configurations, both heuristics show good performances in solving bigger warehouse layout configurations. 
Both heuristics solve 2x2 and 3x3 warehouse layout configurations, which the A* can not solve within the 600-second runtime limit. 
The solution quality of the \texttt{GPT4o:2024-08-06} EoH heuristic is generally better than the \texttt{Qwen2.5:32b-Coder} CEoH heuristic's solution quality for such large warehouse configurations. 
The \texttt{GPT4o:2024-08-06} EoH heuristic shows remarkable performance with 2x2 and 3x3 warehouse configurations with 60\,\% fill percentage with fitness values below 5\,\%.
Both LLM-generated heuristics are unable to find solutions for the most complex instances with 1x1 warehouse layout size and 80\,\% fill percentage.

\section{Conclusion}
\label{sec: Conclusion}
This work investigated the potential of LLMs to generate heuristics for a niche combinatorial optimization problem, the UPMP. We introduced CEoH, an extension of EoH, which enhances heuristic design by incorporating in-context learning through an additional problem-specific description.
\par
Our experiments showed that CEoH consistently outperforms EoH when using the open source models \texttt{Gemma2:27b}, \texttt{Qwen2.5-Coder:32b}, and \texttt{DeepSeekV3:685b}, leading to better heuristic performance and greater robustness against outliers.
In contrast, \texttt{GPT4o:2024-08-06} and \texttt{GPT4o:2024-11-20} found good heuristics with or without additional problem context.
For the \texttt{GPT4o:2024-08-06}, heuristics generated with EoH performed better than those with CEoH, suggesting that too much explicit problem information may sometimes hinder heuristic exploration.
However, the best heuristic, generated by \texttt{Qwen2.5-Coder:32b} with the CEoH framework, achieved an optimality gap of only 4.35\,\% on evaluation instances.
\par
Future research should investigate which types of LLMs benefit the most from additional contextual problem descriptions in heuristic design.
The design of heuristics via LLMs for further niche optimization problems should be studied.

\begin{credits}
\subsubsection{\ackname} This research was funded by the Deutsche Forschungsgemeinschaft
(DFG, German Research Foundation) - project number 276879186 and the European Union - NextGenerationEU - funding code 13IK032.

\end{credits}
%
%
%
\bibliographystyle{splncs04}
\bibliography{sources}


\end{document}